\documentclass{article}
\usepackage[T1]{fontenc}
\usepackage{spconf}
\usepackage{amsmath,amssymb,booktabs,array}
\usepackage[hidelinks]{hyperref}

\hypersetup{pdfauthor={Tongxiao Hu and Junyu Zhou},pdftitle={CROSS-Calib: Selective Channel Calibration for Ultrasound Reconstruction},pdfsubject={Problem analysis, CROSS-Calib method, and experimental validation}}
\usepackage{graphicx}

\usepackage{balance}
\usepackage{placeins}
\title{CROSS-Calib: Selective Channel Calibration for Ultrasound Reconstruction}
\name{\href{https://orcid.org/0009-0002-9805-7461}{Tongxiao Hu}$^{1,*}$ \qquad \href{https://orcid.org/0000-0001-5931-1477}{Junyu Zhou}$^{2,*}$}
\address{$^{1}$University of Oxford \qquad $^{2}$Peking University \\ $^{*}$Corresponding authors.}
\begin{document}
\ninept
\maketitle
\flushbottom
\setlength{\parindent}{0pt}
\begin{abstract}
Ultrasound imaging provides views of internal tissue in real time for medical diagnosis. However, ultrasound signals may be affected by interference, and correcting errors and removing transient interference while preserving valid echoes remains challenging. We propose CROSS-Calib, a channel recovery method that uses predictions from other channels as references for amplitude and phase calibration, while a gate using context selectively combines calibrated measurements with predicted echoes to retain usable samples and repair unreliable ones. On the in vivo test set, CROSS-Calib outperforms all baseline models under gain errors and transient interference, while phantom tests show that the model restores cyst contrast with echo widths of wire targets close to the clean reference. CROSS-Calib can perform channel calibration and selective repair before beamforming to improve echo usability and image quality.
\end{abstract}

\section{Introduction}

Coherent plane wave ultrasound forms images by aligning and superimposing echoes from different receive channels and transmit angles~\cite{montaldo2009}. However, abnormal probe responses and external interference during clinical acquisition may impair image quality. A study of 99 clinical probes classified 31.3\% as defective through sensitivity testing~\cite{vuorenmaa2025}; another clinical image study found that all four defective probes examined reduced the discernibility of structural details~\cite{lorentsson2023}. Transient signal contamination has also been recorded in intraoperative ultrasound, for example, image noise and brief dropout appeared when electrocautery was activated, and the interference disappeared immediately after it stopped~\cite{boat2025}.

Existing reconstruction methods perform optimization at stages of image formation, namely, optimizing the process and restoring channel signals before imaging~\cite{sloun2020review}. For example, adaptive beamforming adjusts the contribution of echoes to each imaging location according to signal covariance or learned receive weights~\cite{synnevag2007,eibmv2010,luijten2020}; INFER jointly estimates scatterer positions, reflectivities, and sound speed~\cite{infer2025}. Deep coherence learning exploits signal coherence across angles for unsupervised beamforming~\cite{dcl2024}, while DP-HAFNet combines adaptive pixel weights and outputs weighted by channel for fusion~\cite{dphafnet2026}. Represented by LISTA proposed by Gregor and LeCun, deep unfolding transforms iterative recovery processes into trainable networks~\cite{lista2010,monga2021}. Mamistvalov and Eldar applied it to undersampled ultrasound reconstruction~\cite{mamistvalov2021}, while Khan et al. applied it to beamforming guided by a model and clutter suppression~\cite{khan2022}.

The latter category processes channel measurements, correcting response errors or repairing corrupted samples before echoes participate in beamforming. Li and Robinson calibrated the timing errors of electronic channels in an array imaging system~\cite{li2012}. Recent work suppresses RF interference from therapeutic ultrasound using 1D U-Nets~\cite{yang2024} and diffusion models~\cite{cai2024}. Learned channel completion uses observed channels to predict missing RF channels in sparse arrays~\cite{vinals2026}. Beamforming combines multichannel echoes, and envelope detection removes complex phase, making it difficult to trace channel errors and their locations of action from images. In contrast, directly processing IQ preserves the angle, channel, and time dimensions, enabling the model to use redundant information from other channels to correct responses, locate contaminated segments, and retain the remaining valid samples. For channels only partially corrupted, the recovery process can use the remaining valid measurements.

The difficulty is that persistent response deviations and local contamination require corrections over different scopes. This paper uses complex gain errors (gain) and local additive interference (burst) to describe these two types of degradation, while also considering missing echoes caused by complete channel failure. Complex gain changes the amplitude and phase of a channel within an acquisition, whereas burst affects only local segments. Therefore, gain correction cannot eliminate local additive interference, while replacement of entire channels may alter uncontaminated echoes. These observations motivate us to consider the estimation of persistent response errors and the repair of local contamination separately.

Our idea is to first use prediction from other channels to calibrate measured echoes, and then control the introduction of predicted values according to local signal features. Accordingly, this paper proposes CROSS-Calib. The unfolded network predicts interleaved groups of withheld channels from the remaining channels, and explicit complex gain fitting produces calibrated IQ. A local gate assigns a weight to each sample, interpolating between the calibrated observation and the predicted reference; channels that cannot be reliably inverted use the reference signal. We synthesize controlled faults on acquired in vivo and phantom IQ and evaluate recovery in terms of signal fidelity, cyst contrast, and wire resolution.

The contributions are:

\begin{itemize}\setlength{\itemsep}{0pt}\setlength{\parsep}{0pt}
\item We propose the CROSS-Calib channel calibration framework, which constructs references from the remaining channels and corrects persistent amplitude and phase errors through explicit complex gain estimation.
\item We design a gate incorporating local context for each sample to interpolate between calibrated observations and predicted references, retain a direct output path for calibrated measured echoes, and control the introduction of values by location.
\item We conduct evaluations under fault types and intensities and examine the role of replacement through ablations sharing the reference and gain estimator.

\end{itemize}

\section{Problem Set}

\subsection{Channel Recovery Task}

This paper addresses channel IQ recovery before beamforming: given multichannel echoes at multiple angles from the same acquisition, estimate the signals before abnormal responses, local interference, or channel failure occurred. The recovered result should preserve the amplitude, phase, and temporal structure of valid echoes for subsequent coherent superposition. Let $S,Y\in\mathbb{C}^{A\times C\times T}$ denote the target IQ and corrupted observation, respectively, where $A$, $C$, and $T$ denote the number of transmit angles, receive channels, and echo samples, respectively. The observation relationship is

\begin{equation}
\begin{aligned}
Y_{act}&=m_cg_cS_{act}+B_{act},\\
g_c&=e^{\alpha_c+\mathrm{j}\phi_c}.
\end{aligned}
\end{equation}

Here, $m_c\in\{0,1\}$ denotes channel availability; $\alpha_c$ and $\phi_c$ denote the log amplitude deviation and phase deviation, respectively, and the complex gain $g_c$ remains constant across angles and time within an acquisition; $B$ denotes additive interference acting on local segments. The restorer receives only $Y$ and outputs $\hat S$ without knowledge of fault labels, locations, or error parameters, followed by delay and sum (DAS) beamforming.

\subsection{Observations and Phenomena}

\noindent\textbf{Observation 1: Erroneous observations may become their own calibration reference.} Let $V$ be the predicted reference. If the predictor copies the target observation so that $V=Y$, fitting $Y\approx\hat gV$ on nonzero signals will yield $\hat g=1$, and response errors already present in the input cannot be identified.

\noindent\textbf{Observation 2: gain and burst impose different requirements on recovery.} On the 1,825 V4 acquisitions, using three input angles and the heavy fault injection protocol, RF-inpainting and the unfolded model containing explicit gain fitting and analytical residual suppression have median IQ NMSE 0.01747 and 0.003570 under gain, respectively, but 0.0002332 and 0.003045 under burst, reversing their ranking.

\noindent\textbf{Observation 3: Differences between observations and references are not entirely contamination.} For available channels and nonzero $\hat g$, let $U=Y/\hat g$; omitting subscripts gives

\begin{equation}
U-V=\frac{B}{\hat g}
 +\left(\frac{g}{\hat g}-1\right)S
 +(S-V).
\end{equation}

The residual mixes calibrated interference, remaining gain errors, and reference prediction errors. Even when $B=0$ and $\hat g=g$, $S-V$ remains; pulling the output toward the reference in this case may instead move it away from valid measurements.

\noindent\textbf{Principle 1: Construct references from other channels.}

\noindent\textbf{Principle 2: Allocate estimation and repair according to the scope of the errors.}

\noindent\textbf{Principle 3: Repair selectively and retain valid measurements.}

\begin{figure}[t]
\centering
\includegraphics[width=\columnwidth,trim=11bp 0bp 18bp 0bp,clip]{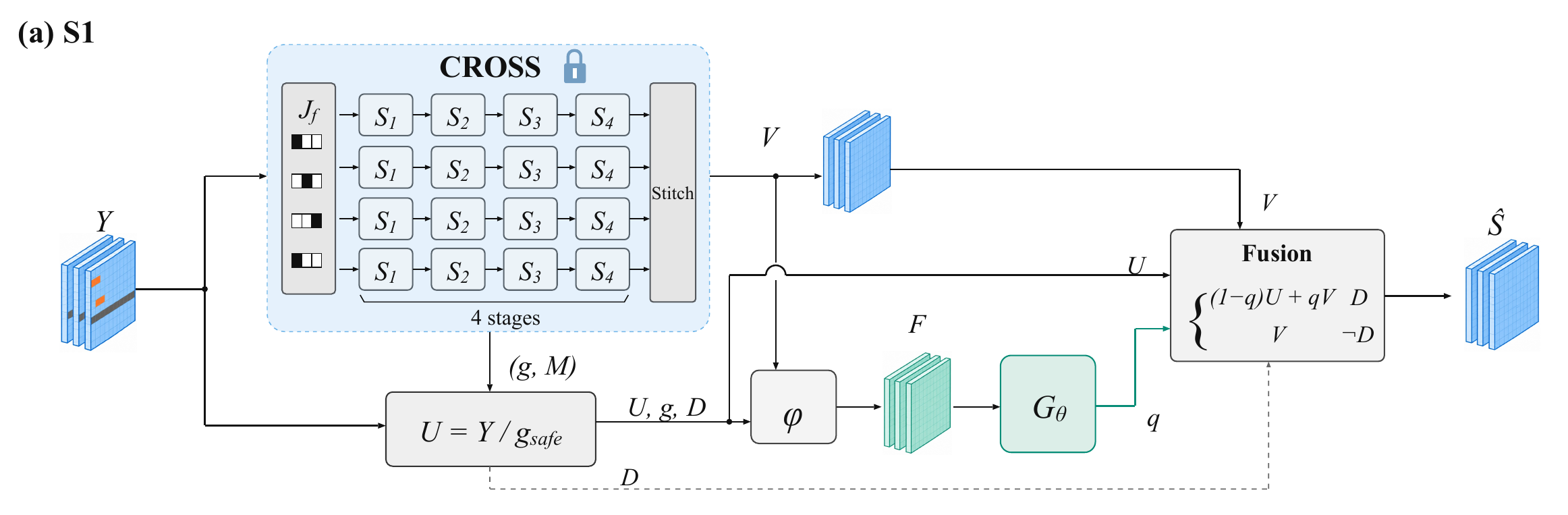}\par\vspace{2pt}
\includegraphics[width=.637\columnwidth,trim=7bp 0bp 2bp 0bp,clip]{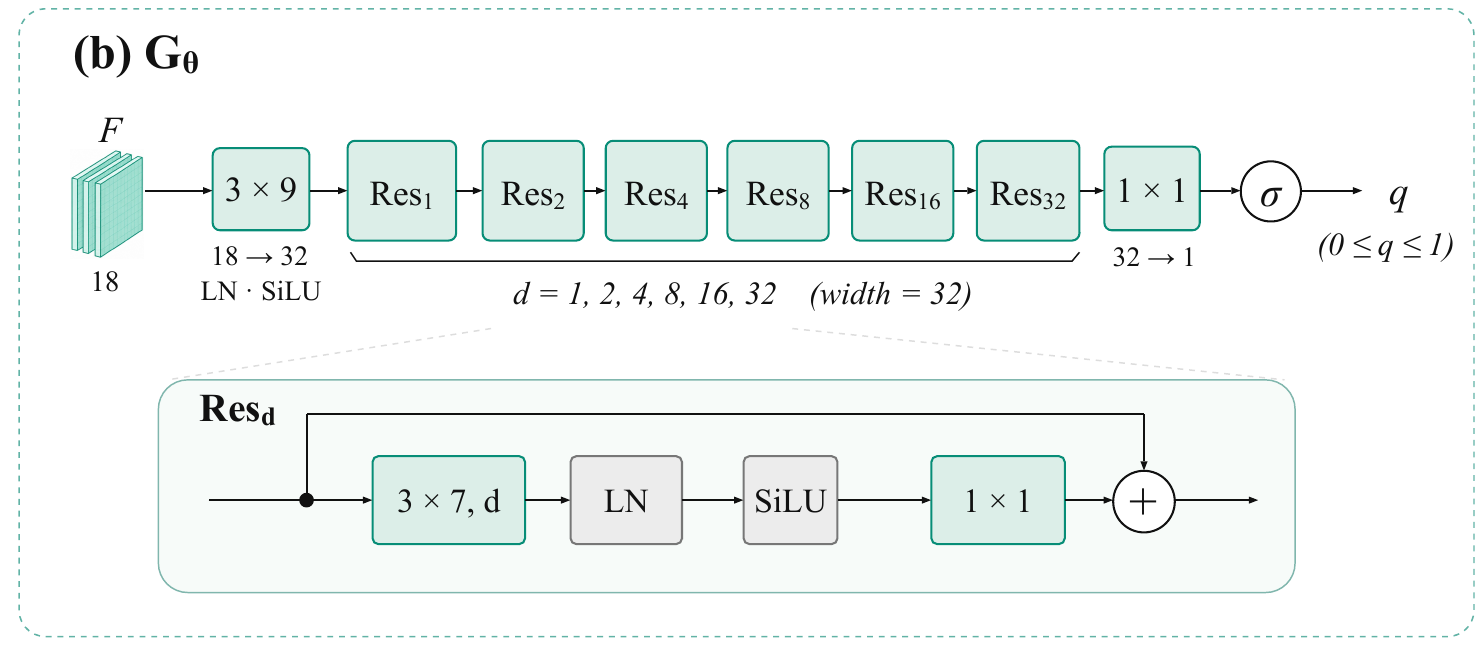}\hfill
\includegraphics[width=.359\columnwidth,trim=2bp 0bp 6bp 0bp,clip]{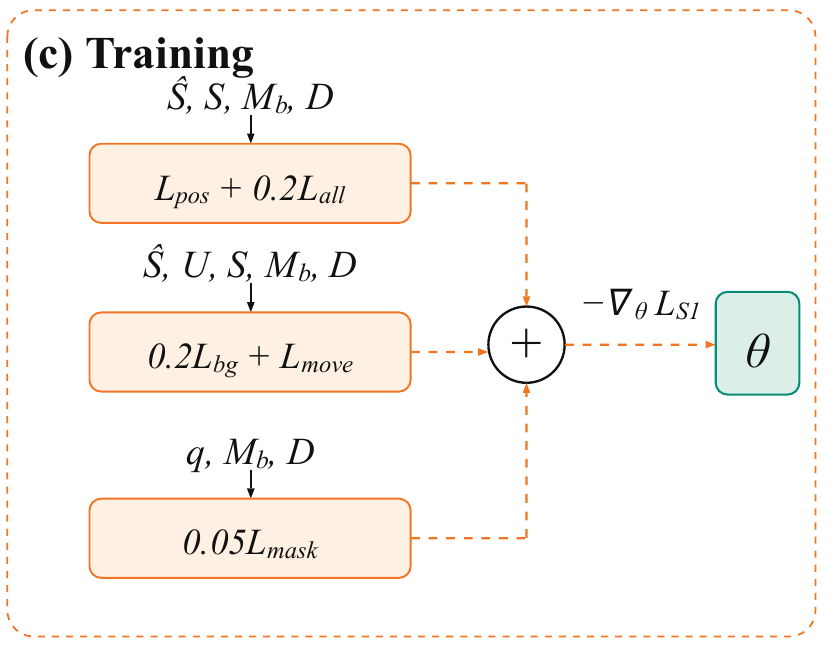}
\caption{CROSS-Calib: (a) workflow; (b) local gate; (c) gate training.}
\label{fig:architecture}
\end{figure}

\section{Method}

Given the multichannel observation at multiple angles $Y$, the reference network generates candidate echoes $V$ from other channels, and the gain estimator estimates $\hat g$ to obtain the calibrated observation $U=Y/\hat g$. The local gate $q\in[0,1]^{A\times C\times T}$ determines how much predicted information to introduce at each location, with the output given by
\begin{equation}
 \hat S_{act}=\begin{cases}
 (1-q_{act})U_{act}+q_{act}V_{act},&d_c=1,\\
 V_{act},&d_c=0,
 \end{cases}
\label{eq:m_output}
\end{equation}
Here, $d_c$ indicates that the observation is available and its estimated gain can be reliably inverted. $\hat S$ is imaged by DAS. 

\subsection{References across Channels and Unfolded Recovery}

The reference network uses an unfolded architecture with four stages, with a residual U-Net~\cite{unet2015,resnet2016} shared across stages that takes the real and imaginary IQ components and the availability mask as input. Receive channels are divided into four interleaved groups by channel index. Each time, all angle and time samples of one group are masked, and only the other three groups are used to predict its echoes; the masked group is termed the ``withheld group.'' After each has served as the withheld group, the predicted echoes are stitched at their original channel positions to form the full reference from other channels $V$.

Let the withheld group be $J_f=\{c:c\bmod4=f\}$; only the remaining channels are used to estimate the availability mask $m_f$ and RMS normalization scale $s_f$. Let $y_f=m_f\odot Y/s_f$; the echoes, gains, and auxiliary additive terms within the branch are initialized with $(x_f^0,g_f^0,b_f^0)=(y_f,1,0)$.

At each stage, a shared residual U-Net generates $z_f^k=P_\psi(x_f^k,m_f,k)$, and combines candidates with measurements:
\begin{equation}
 x_f^{k+1}=\frac{m_f\odot(g_f^k)^*\odot(y_f-b_f^k)+\rho_kz_f^k}
 {m_f\odot|g_f^k|^2+\rho_k}.
\label{eq:m_update}
\end{equation}
Eq.~\eqref{eq:m_update} is the solution in closed form of a weighted quadratic subproblem involving measurement error and deviation from the candidate; $\rho_k>0$ controls the prior weight, $*$ denotes complex conjugation, and channel quantities are broadcast over the angle and time dimensions. Subsequently, $g_f$ is updated through complex least squares and shrinkage centered at unit gain, and $b_f$ is updated through local residual shrinkage.

After restoring the predictions from four stages to the original IQ scale, stitching the predicted groups at their channel positions is written as $V_{:,J_f,:}=s_f(x_f^4)_{:,J_f,:}$.

\begin{table}[t]
\centering
\fontsize{9}{11}\selectfont
\caption{V8 DAS SSIM $\uparrow$ (median, $n=1605$).}
\label{tab:baselines}
\setlength{\tabcolsep}{2.5pt}
\begin{tabular}{lrrrr}
\toprule
&\multicolumn{2}{c}{Gain}&\multicolumn{2}{c}{Burst}\\
Method&A3&A11&A3&A11\\
\midrule\multicolumn{5}{c}{Heavy}\\
No repair&0.6873&0.7426&0.8884&0.9595\\
RF-inpainting&0.9083&0.9315&0.9958&0.9989\\
IRLS-CG&0.6627&0.7192&0.8363&0.8667\\
Detect--mask CG&0.5090&0.5577&0.6562&0.6812\\
J-Sense-US$^\dagger$&0.6938&0.7452&0.8074&0.8480\\
DP-HAFNet-DAS$^\dagger$&0.6811&0.7150&0.8372&0.8802\\
HIFU-Diff-IQ&0.7267&0.7682&0.7438&0.8080\\
CROSS-Calib&\textbf{0.9991}&\textbf{0.9996}&\textbf{0.9977}&\textbf{0.9995}\\
\midrule\multicolumn{5}{c}{Extreme}\\
No repair&0.1961&0.2498&0.2900&0.5452\\
RF-inpainting&0.5329&0.6049&0.9361&0.9850\\
IRLS-CG&0.2720&0.3549&0.7486&0.8471\\
Detect--mask CG&0.1533&0.1918&0.5898&0.6697\\
J-Sense-US$^\dagger$&0.4856&0.5591&0.6192&0.8133\\
DP-HAFNet-DAS$^\dagger$&0.2179&0.2712&0.3186&0.5455\\
HIFU-Diff-IQ&0.3039&0.3672&0.2122&0.3508\\
CROSS-Calib&\textbf{0.9096}&\textbf{0.9500}&\textbf{0.9871}&\textbf{0.9970}\\
\bottomrule
\end{tabular}
\end{table}

\begin{table}[t]
\centering
\fontsize{9}{11}\selectfont
\caption{V8 A11 DAS SSIM $\uparrow$ (heavy/extreme), $n=1605$.}
\label{tab:ablation}
\setlength{\tabcolsep}{1.2pt}
\begin{tabular}{lccc}
\toprule
Fault&\shortstack{Analytical\\readout}&\shortstack{Gain correction\\alone}&CROSS-Calib\\
\midrule
none&0.9998/0.9998&\textbf{1.0000}/\textbf{1.0000}&\textbf{1.0000}/\textbf{1.0000}\\
dead&0.9353/0.7883&\textbf{0.9394}/\textbf{0.7969}&\textbf{0.9394}/\textbf{0.7969}\\
gain&0.9977/0.9355&\textbf{0.9996}/0.9471&\textbf{0.9996}/\textbf{0.9500}\\
burst&0.9845/0.8799&0.9595/0.5452&\textbf{0.9995}/\textbf{0.9970}\\
\bottomrule
\end{tabular}
\end{table}

\begin{table*}[t]
\centering
\caption{054GS A3 metrics (heavy/extreme). Best repairs are bold; FWHM closest to the clean reference is best.}
\label{tab:phantom}
\fontsize{9}{11}\selectfont
\setlength{\tabcolsep}{3pt}
\begin{tabular}{llrrrrr}
\toprule
Fault & Method & gCNR & CNR (dB) & CR (dB) & Lateral FWHM (mm) & Axial FWHM (mm)\\
\midrule
-- & Clean reference & 0.929 & 3.90 & 19.95 & 0.515 & 0.408\\
\midrule
gain & No repair & 0.791/0.306 & 2.38/-6.08 & 12.69/3.42 & 0.514/0.522 & 0.408/0.408\\
 & RF-inpainting & 0.871/0.599& 3.22/0.15& 15.88/8.29& 0.516/0.524& 0.407/0.407\\
 & IRLS-CG & 0.798/0.351& 2.49/-4.61& 12.87/4.04& 0.495/0.499& 0.427/0.422\\
 & Detect--mask CG & 0.791/0.307& 2.38/-6.09& 12.69/3.41& 0.534/0.526& 0.452/0.459\\
 & J-Sense-US & 0.830/0.662& 2.86/0.96& 14.07/9.37& 0.495/0.497& 0.428/0.427\\
 & DP-HAFNet-DAS & 0.810/0.321& 2.77/-5.44& 11.35/3.04& 0.519/0.533& 0.411/0.414\\
 & HIFU-Diff-IQ & 0.525/0.236& -1.05/-10.72& 6.87/1.86& 0.527/0.524 & 0.464/0.463\\
 & \textbf{CROSS-Calib} & \textbf{0.930}/\textbf{0.902}& \textbf{3.89}/\textbf{3.64}& \textbf{19.92}/\textbf{17.65}& \textbf{0.515}/\textbf{0.515}& \textbf{0.408}/\textbf{0.408}\\
\midrule
burst & No repair & 0.871/0.397 & 3.42/-3.72 & 17.31/4.83 & 0.515/0.511 & 0.408/0.406\\
 & RF-inpainting & 0.927/0.899& 3.87/3.56& 19.64/17.71& \textbf{0.516}/\textbf{0.517}& \textbf{0.408}/\textbf{0.408}\\
 & IRLS-CG & 0.926/0.860& 3.83/3.15& 19.31/15.09& 0.497/0.500& 0.425/0.417\\
 & Detect--mask CG & \textbf{0.928}/0.911& 3.88/3.68& \textbf{19.83}/18.48& 0.538/0.524& 0.451/0.457\\
 & J-Sense-US & 0.907/0.861& 3.69/3.17& 17.97/15.19& 0.490/0.501& 0.428/0.420\\
 & DP-HAFNet-DAS & 0.903/0.455& \textbf{3.89}/-1.91& 14.47/4.03& 0.522/0.537& 0.412/0.413\\
 & HIFU-Diff-IQ & 0.645/0.220& 0.52/-15.41& 9.05/1.01& 0.543/0.541 & 0.411/0.413\\
 & \textbf{CROSS-Calib} & 0.926/\textbf{0.916}& 3.84/\textbf{3.73}& 19.68/\textbf{18.92}& \textbf{0.516}/\textbf{0.517}& \textbf{0.408}/\textbf{0.408}\\
\bottomrule
\end{tabular}
\end{table*}

\subsection{Persistent Complex Gain Calibration}

With $V$ fixed, availability $\hat m$ is estimated from the observation, the residual is initialized with $b^{(0)}=0$, and the $\ell$th iteration computes
\begin{equation}
 r_c^{(\ell)}=\frac{\sum_{a,t}\hat m_cV_{act}^*(Y_{act}-b_{act}^{(\ell)})}
 {\max(\sum_{a,t}\hat m_c|V_{act}|^2,\epsilon)}.
\label{eq:m_gain}
\end{equation}
Fits close to unit gain are kept at 1, and channels with low reference energy also use unit gain to avoid introducing amplitude and phase corrections when evidence is insufficient.

To reduce the influence of local contamination on fitting over the entire channel, after the gain update, residual shrinkage based on local RMS and background intensity is applied to $Y-gV$, serving as the auxiliary additive term for the next fit. Two iterations yield $\hat g$, after which $U=Y/\hat g$ is computed.

\subsection{Selective Repair Using Context}

We compare $U$, $V$, and $D=U-V$ in the calibrated domain, so that persistent amplitude and phase deviations undergo correction before the network assesses the reliability of local measurements.

The ratio of local RMS to the median RMS of surrounding windows describes abrupt changes in a segment relative to its background. The local normalized correlation between $U$ and $V$ describes whether waveforms are consistent. These quantities, together with the real and imaginary IQ components, gain, and validity information, form features $\Phi$.

A lightweight convolutional network produces $q=\sigma(G_\theta(\Phi))$. The network extracts local structure along channels and time, expands the context through temporally dilated convolutions, and shares weights across angles. The output is initialized close to $q=0$.

\subsection{Training Objective}

First, the backbone is pretrained using paired IQ before and after fault injection. The backbone and gain estimator are fixed, and only the gate parameters $\theta$ are updated. $V$ and $\hat g$ are recomputed per input.

Let the burst support known during training be $M$ and the true mask of valid channels be $m^\star$. Define the contaminated region $P=d\cap m^\star\cap M$ and the remaining valid region $H=d\cap m^\star\cap\neg M$. With $\langle\cdot\rangle_\Omega$ denoting the regional mean and $e$ denoting the average total target energy per valid channel, we use
\begin{equation}
\begin{aligned}
 L_P&=\langle|\hat S-S|^2/e\rangle_P,\quad
 L_H=\langle|\hat S-S|^2/e\rangle_H,\\
 L_{\rm keep}&=\langle|\hat S-U|^2/e\rangle_H,\\
 L_{\rm mask}&=-\langle\log q\rangle_P-\langle\log(1-q)\rangle_H,\\
 L_{\rm all}&=\|\hat S-S\|_F^2/\|S\|_F^2.
\end{aligned}
\label{eq:m_losses}
\end{equation}
The total loss is
\begin{equation}
 L=L_P+0.2L_H+L_{\rm keep}+0.05L_{\rm mask}+0.2L_{\rm all}.
\label{eq:m_total}
\end{equation}

\section{Experiments}
\begin{figure}[t]
\centering
\includegraphics[width=\columnwidth]{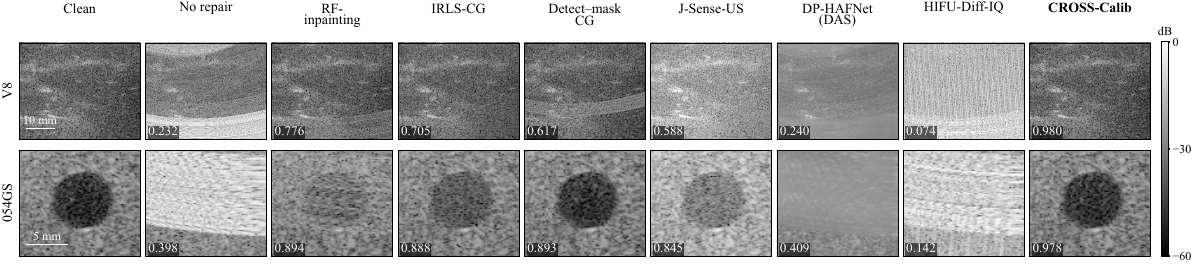}
\caption{Extreme A3 burst repair on V8 and 054GS. 60 dB display; values are DAS SSIM.}
\label{fig:cyst}
\end{figure}

\subsection{Dataset and Experimental Settings}
The EPFL ultrafast ultrasound dataset contains approximately 20,000 in vivo RF acquisitions from nine healthy volunteers~\cite{vinals2026}. Each IQ input has 192 receive channels and 2,133 samples per channel, with three or eleven transmit angles (A3/A11). We use 13,976 acquisitions from V1/V2/V3/V6/V7/V9 for training, 1,825 from V4 for validation and model selection, 1,605 from V8 for in vivo testing, and 1,013 from 054GS for phantom testing. Each acquisition provides A3 and A11 inputs, with its IQ before injection as the target.

We separately inject complex gain abnormalities (gain), local additive interference (burst), and channel failure (dead). Under heavy, gain applies amplitude deviations of $\pm6$~dB and phase deviations of $\pm60^\circ$ to 16 random channels, with independently chosen signs. dead zeros 16 random channels across all angles and samples. burst adds eight Gaussian noise segments of 128 samples per input, with RMS 20~dB above the corresponding signal segment. extreme increases the gain/dead channel count to 48, gain deviations to $\pm12$~dB and $\pm90^\circ$, and burst to 32 segments of 256 samples at +25~dB.

The gate is trained for ten FP32 epochs on four RTX 5090 GPUs, with a burst/gain/none/dead ratio of $4{:}3{:}2{:}1$ and initial AdamW~\cite{adamw2019} learning rate $3\times10^{-4}$. Baselines are RF-inpainting~\cite{vinals2026}, Huber IRLS-CG~\cite{huber1964,hestenes1952}, detect--mask CG~\cite{rousseeuw1993,hestenes1952}, J-Sense-US adapted to ultrasound~\cite{deepjsense2021}, DP-HAFNet~\cite{dphafnet2026}, and HIFU-Diff-IQ~\cite{cai2024}. The latter adapts HIFU-Diff to real/imaginary IQ channels, uses the same training split and fault mixture for four BF16 epochs, selects the checkpoint with the lowest training loss, and uses 20 DDIM steps ($\eta=0.5$). Normalization at test time uses only the corrupted input scale, restored after prediction; each case uses one stochastic sample. All methods receive identical test faults. IQ NMSE uses no amplitude/phase alignment; DAS SSIM~\cite{ssim2004} uses a 60~dB display range and the same acquisition's clean image at the same angle count. Tables report case medians; paired in vivo values denote A3/A11.

\subsection{Test Set Data across Fault Types and Intensities}
CROSS-Calib has the highest SSIM in all eight gain/burst, heavy/extreme, and A3/A11 settings in Table~\ref{tab:baselines}; RF-inpainting remains the strongest baseline. Under extreme gain, CROSS-Calib reaches 0.9096/0.9500, versus 0.5329/0.6049 for RF-inpainting.

Under heavy burst, RF-inpainting reaches 0.9958/0.9989 SSIM, close to CROSS-Calib. Under extreme burst, the gap widens: CROSS-Calib reaches 0.9871/0.9970 versus 0.9361/0.9850. Relative IQ NMSE reductions over RF-inpainting are 74.1\%/82.4\% for heavy burst and 88.6\%/91.6\% for extreme burst; for heavy and extreme gain, they are 99.2\%/99.4\% and 91.6\%/94.3\%. SSIM near its ceiling under heavy burst leaves little room for improvement, whereas IQ NMSE remains sensitive to residual channel errors. The HIFU-Diff-IQ adaptation reaches 0.7438/0.8080 under heavy burst and 0.2122/0.3508 under extreme burst. Its SSIM of 0.9216/0.9359 without faults also shows changes to valid echoes, supporting the need to preserve usable measurements during repair.

\subsection{Cyst Contrast and Wire Target Profiles}
On 196 hypoechoic cyst acquisitions (Table~\ref{tab:phantom}), CROSS-Calib's A3 heavy/extreme gain gCNR~\cite{gcnr2020} is 0.930/0.902, close to the clean reference of 0.929. Under extreme gain, RF-inpainting reaches 0.599, J-Sense-US reaches 0.662, and HIFU-Diff-IQ reaches 0.236. CROSS-Calib gives the strongest CNR/CR recovery: 3.64/17.65~dB versus the clean reference's 3.90/19.95~dB. CNR is $20\log_{10}(|\mu_i-\mu_o|/\sqrt{\sigma_i^2+\sigma_o^2})$ and CR is $20\log_{10}(\mu_o/\mu_i)$, where $i/o$ denote cyst/background envelope statistics.

Heavy burst yields similar gCNR: CROSS-Calib reaches 0.926, RF-inpainting 0.927, and detect--mask CG 0.928. Under extreme burst, CROSS-Calib leads at 0.916, followed by detect--mask CG at 0.911 and RF-inpainting at 0.899; its CNR and CR lead at 3.73 and 18.92~dB. Fig.~\ref{fig:cyst} illustrates these effects. Lateral/axial FWHM is 0.517/0.408~mm, versus 0.515/0.408~mm for the reference, indicating contrast recovery with nearly unchanged widths of wire targets.

\subsection{Ablation Study}
Table~\ref{tab:ablation} compares output rules on 1,605 V8 acquisitions using A11 input. The variants share reference and gain estimates: gain correction alone removes local repair, analytical readout uses deterministic residual suppression, and CROSS-Calib uses learned gating.

Under heavy/extreme burst, CROSS-Calib reaches 0.9995/0.9970 SSIM, compared with 0.9595/0.5452 for gain correction alone and 0.9845/0.8799 for analytical readout. This supports the learned gate's contribution to local interference repair. Under extreme gain, CROSS-Calib improves SSIM from 0.9471 to 0.9500 relative to gain correction alone. For none and dead, the two variants match at the reported precision. Dead channels use the shared predicted reference directly, so gating adds no reconstruction capability to those channels with missing echoes.

\section{Conclusion}
This paper proposes CROSS-Calib, combining references from other channels, explicit complex gain calibration, and a local gate before beamforming to address persistent response deviations and local contamination separately. In experiments with heavy/extreme controlled faults on acquired in vivo and phantom IQ, the method improves signal fidelity and imaging quality under gain and burst conditions; ablations with the reference and gain estimator fixed show that the local gate can reduce residual burst errors while retaining the benefits of gain calibration.

\par\penalty-10000
\clearpage
{
\bibliographystyle{IEEEbib}
\bibliography{references}
}
\end{document}